\documentclass[conference]{IEEEtran}
\IEEEoverridecommandlockouts
\usepackage{cite}
\usepackage{amsmath,amssymb,amsfonts}
\usepackage{algorithmic}
\usepackage{graphicx}
\usepackage{textcomp}
\usepackage{xcolor}
\usepackage{booktabs}

\def\BibTeX{{\rm B\kern-.05em{\sc i\kern-.025em b}\kern-.08em
    T\kern-.1667em\lower.7ex\hbox{E}\kern-.125emX}}
\begin{document}

\title{How Reliable AI Chatbots are for Disease Prediction from Patient Complaints?\\

}

\author{
\IEEEauthorblockN{Ayesha Siddika Nipu}
\IEEEauthorblockA{\textit{Computer Science \& Software Engineering} \\
\textit{University of Wisconsin-Platteville}\\
Platteville, WI, USA \\
nipua@uwplatt.edu}
\and
\IEEEauthorblockN{K M Sajjadul Islam}
\IEEEauthorblockA{\textit{Computer Science} \\
\textit{Marquette University}\\
Milwaukee, WI, USA \\
sajjad.islam@marquette.edu}
\and
\IEEEauthorblockN{Praveen Madiraju}
\IEEEauthorblockA{\textit{Computer Science} \\
\textit{Marquette University}\\
Milwaukee, WI, USA  \\
praveen.madiraju@marquette.edu}
}

\maketitle

\begin{abstract}
Artificial Intelligence (AI) chatbots leveraging Large Language Models (LLMs) are gaining traction in healthcare for their potential to automate patient interactions and aid clinical decision-making. 
This study examines the reliability of AI chatbots, specifically GPT 4.0, Claude 3 Opus, and Gemini Ultra 1.0, in predicting diseases from patient complaints in the emergency department.
The methodology includes few-shot learning techniques to evaluate the chatbots' effectiveness in disease prediction. We also fine-tune the transformer-based model BERT and compare its performance with the AI chatbots. 
Results suggest that GPT 4.0 achieves high accuracy with increased few-shot data, while Gemini Ultra 1.0 performs well with fewer examples, and Claude 3 Opus maintains consistent performance. 
BERT's performance, however, is lower than all the chatbots, indicating limitations due to limited labeled data.
Despite the chatbots' varying accuracy, none of them are sufficiently reliable for critical medical decision-making, underscoring the need for rigorous validation and human oversight.
This study reflects that while AI chatbots have potential in healthcare, they should complement, not replace, human expertise to ensure patient safety.
Further refinement and research are needed to improve AI-based healthcare applications' reliability for disease prediction.

\end{abstract}


\section{Introduction}

Artificial Intelligence (AI) chatbots are software programs that use artificial intelligence to imitate human interaction. 
It has the ability to engage with humans through text or speech, frequently utilizing natural language processing (NLP) to comprehend and address inquiries. 
AI chatbots are utilized in customer service, virtual assistants, and other interactive platforms to deliver automated responses \cite{bostrom2018ethics}. 
Large Language Models (LLMs), such as OpenAI's GPT or Google's BERT, are specific types of AI models that employ transformer-based architectures to analyze vast quantities of multimodal data. 
These models produce content that resembles human writing, as they can comprehend the meaning and connections between words and phrases \cite{ouyang2022training}. 
This makes them well-suited for jobs like generating text, translating languages, and summarizing information. 
LLMs, when linked with AI chatbots, enhance the chatbot's capabilities by enabling advanced and contextually aware interactions. 
This expansion of functionality makes the chatbots more adaptable and applicable to a wide range of fields, including healthcare.

AI chatbots are becoming increasingly prominent in healthcare, especially after the integration of LLMs that can respond to free-text queries without specific task training. 
These chatbots offer 24/7 health advice and support, potentially minimizing the need for in-person consultations and improving patient outcomes. They can also provide valuable data and insights to healthcare professionals, supporting more informed decision-making \cite{nadarzynski2019acceptability}.
However, AI lacks the capacity for empathy, intuition, and the extensive experience inherent in medical professionals \cite{altamimi2023artificial}. 
These human traits are critical for effective patient care, especially when interpreting subtle language nuances and non-verbal cues. 
AI chatbots operate based on pre-set data and algorithms, which limits the quality of their recommendations to the data they process. 
If this data is incomplete or biased, it can lead to inaccurate or potentially harmful outcomes. 
This underscores the importance of validating AI chatbot outputs in healthcare to ensure patient safety and care quality.

Electronic Health Record (EHR) is a digital version of a patient's medical history, maintained by healthcare providers.
Patient complaints, also known as chief complaints (CC), are integral to EHRs, serving as brief statements explaining why a patient is seeking medical care. 
It contains freeform text, usually made up of one or more incomplete structured sentences.
It is typically recorded when a patient registers at a clinic or emergency department and may also be documented by clinicians in various medical notes during a patient's care, such as progress notes, discharge notes, and transfer notes \cite{islam2023autocompletion}. 
Disease prediction in CC is important because it helps streamline the medical triage process by categorizing CCs into distinct groups. 
Accurate classification also aids in identifying patterns and trends in patient complaints, contributing to better resource planning and improved patient outcomes. 
In this study, our aim is twofold:
\begin{itemize}
    \item Predict the presence of disease by analyzing patient complaints, leveraging three widely used general-purpose AI-based chatbots (GPT 4.0, Claude 3 Opus, Gemini Ultra 1.0), employing few-shot learning techniques.
    \item Assess the effectiveness and reliability of AI chatbots in predicting disease from patient complaints, providing insights into the potential role of AI in medical diagnostics.
\end{itemize}

\section{Background Study}
\subsection{Chief Complaint Classification in Electronic Health Record}
Numerous recent studies have focused on CC, emphasizing syndromic surveillance \cite{conway2013using}, automated CC classification \cite{lee2019chief, sorensen2021predicting, li2019improving}. CC classification is foundational in the patient care journey, affecting individual patient outcomes and broader healthcare system efficiency. Techniques used in CC classification support various tasks, including hospital admission prediction, symptom-based risk stratification, and so on. For instance, Lee et al. \cite{lee2019chief} compare recurrent neural network models (LSTM and GRU) with traditional bag-of-words classifiers for automated classification of syndromic surveillance data in emergency department records. Sørensen et al. \cite{sorensen2021predicting} investigate 30-day mortality and readmission rates among emergency department patients based on CCs, using logistic regression to determine crude and adjusted odds ratios. Wang et al. \cite{li2019improving} introduce a rare disease classification algorithm combining ``bag of words" with ``bag of knowledge terms" derived from a knowledge graph (KG) to enhance the accuracy of symptom-based rare disease detection. These studies highlight the evolving role of CCs in healthcare, demonstrating the potential to improve patient outcomes, resource management, and disease detection.

\subsection{AI Chatbots in Healthcare}
AI chatbots are transforming healthcare by enhancing patient-specific interactions and supporting clinical practices. Recent studies illustrate the diverse applications of AI chatbots, demonstrating their potential in areas such as patient question answering (QA), hearing health care, and medical education. For example, Hamidi et al.\cite{hamidi2023evaluation} evaluate AI chatbots for patient-specific QA from clinical notes, comparing the accuracy and relevance of responses from LLMs like ChatGPT, Bard (now Gemini), and Claude. Swanepoel et al.\cite{swanepoel2023rise} explores the revolutionary impact of AI chatbots in hearing health care, highlighting how they can engage in human-like conversational responses. In medical education, Ghorashi et al.\cite{ghorashi2023ai} discuss the use of AI-powered chatbots to simplify complex concepts and serve as interactive tutors while underscoring the importance of referencing evidence-based resources. These studies collectively indicate the growing importance of AI chatbots and their potential to improve patient care in the healthcare system.

\subsection{Recent Trends in Text Classification}
Recent trends in text classification have shifted from traditional statistical models to more sophisticated transformer-based models, largely due to the advent of LLMs like BERT, GPT, and Llama. These models offer enhanced accuracy, scalability, and adaptability, allowing for efficient handling of large datasets and fine-tuning for various tasks. For example, Pal et al.\cite{pal2017machine} explore the use of machine learning in the stylometric analysis of literature, illustrating the application of traditional statistical models. Fields et al.\cite{fields2024survey} examine the growing popularity of LLMs for text classification, emphasizing their role in improving accuracy and scalability while raising important ethical considerations. Ge et al. \cite{ge2023few} highlight the use of few-shot learning (FSL) in medical text classification, indicating its usefulness given the limited labeled datasets in the medical field. These examples reflect the evolving landscape of text classification and the increasing role of advanced AI models.

 \begin{figure}[tbp]
\centerline{\includegraphics[width=0.48\textwidth]{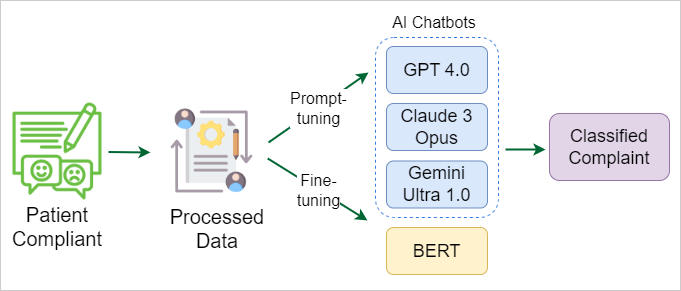}}
\caption{Process Flow of Disease Prediction}
\label{Fig:ProcessFlow}
\end{figure}

\section{Methodology}
Our proposed experiment comprises four major steps: data preprocessing, prompt tuning, fine-tuning, and classifying the complaints. Figure~\ref{Fig:ProcessFlow} provides a visual representation of the process flow, detailing the sequential steps involved in predicting gout flares from CC. 

\subsection{Dataset Description}\label{AA}
In our experiment, we use the Gout Emergency Department Chief Complaint Corpora (GED3C), which comprises two distinct datasets: GOUT-CC-2019-CORPUS and GOUT-CC-2020-CORPUS \cite{osborne2020gout}. 
These corpora contain free-text CCs from an academic medical center in the Deep South, predominantly written by triage nurses in an urban setting. 
The smaller GOUT-CC-2019-CORPUS includes 300 CCs from 2019 specifically selected for predicting the presence of ``gout" in the CC. 
The advantage of using this dataset is that these are annotated retrospectively for predicted gout flare status based on manual review of CCs, with a subset undergoing chart review by rheumatologists to verify gout flare status. 
The CCs are further processed for de-identification to meet Health Insurance Portability and Accountability Act (HIPAA) specifications, with personal health information (PHI) removed and replaced with HIPAA classes of protected information. 
Additionally, time-specific information that could lead to patient identification is also eliminated. 
Each row in the dataset represents a CC and contains three fields: the `Chief Complaint' text, `Predict' an analysis by the authors of the dataset to identify whether the complaint may be related to a gout flare, and `Consensus' on gout flare status determined through expert review. 
The `Consensus' column contains yes (Y), no (N), unknown (U), or unmarked (-) labels.
Table~\ref{Table:Dataset} illustrates a few samples from our selected dataset.

\subsection{Dataset Preprocessing}
The CC column contains free text, typically consisting of one or more incomplete or improperly structured sentences. 
It is often written in abbreviated forms and rich in medical acronyms. 
We identify that a CC could be divided into two parts: the first part captures a patient's complaint about their current health condition, while the second part pertains to their past medical or personal history \cite{islam2023autocompletion}. 
To clean these texts, we focus on medical acronyms indicating past medical or personal history, such as PMH, PMHX, HX, PSHX, SHX, and FHX. 
We create a custom function to split the CC into two parts, keeping only the initial complaint while removing any content related to past medical or personal history. 
This division is critical to separate the current complaint from irrelevant or potentially sensitive information. 
We use the Python NLP library Stanza to help separate these sentences for easier processing.

To further explore the dataset, we perform lemmatization and find a total of 670 unique words. Table~\ref{Table:wordFrequency} shows some of the most and least frequent words from our selected dataset.
We then group the dataset based on the `Consensus' and identify 118 documents labeled as N, 103 as `-', 70 as Y, and 9 as U. 
To ensure the integrity and usability of our dataset for subsequent analyses, we decide to remove the unmarked patient complaints. 
This step is essential to maintaining the quality and relevance of the data, focusing only on entries that have definitive and informative labels provided by healthcare experts. 
We consider the proportions of the `Consensus' values for train-test-split to preserve the relative frequency of each category, which serves as the basis for stratified sampling. 
Stratified sampling involves selecting samples from each category according to their proportions, ensuring representativeness across the dataset. 
This approach helps create a balanced subset that reflects the original distribution, avoiding biases \cite{neyman1992two}.

\begin{table}[tb]
\caption{Chief Complaint Dataset Examples}
\begin{center}
\begin{tabular}{p{6cm}c}
\toprule
\textbf{Chief Complaint} & \textbf{\textit{Consensus}}\\
\hline right shoulder pain/redness/swelling x 3 days, pmh ESRD, HTN, CVA, gouty and osteoarthritis
 & U \\
\hline
abd pain, NV x 1 year, worse over last 7 months. has lost 100lbs in last year. pmh: hernia repair \textless\textless DATE\textgreater\textgreater, gout, migraines & N \\
\hline
Bilateral ankle and foot pain x 2 weeks. C/o chills, coughing x 2 weeks. Seen PCP, dx with athletes foot to left foot. PMH gout, HTN & Y \\
\bottomrule
\end{tabular}
\label{Table:Dataset}
\end{center}
\end{table}

\subsection{AI Chatbots}
In this study, we explore the emerging application of LLMs in the medical domain by evaluating the performance of three leading LLMs: GPT 4.0, Claude 3 Opus, and Gemini Ultra 1.0 — on our patient feedback dataset. 
Our study provides a deeper understanding of the strengths and limitations of each LLM when applied to medical data, offering insights into the future potential of artificial general intelligence in healthcare.

\begin{table}[tbp]
  \caption{Word Frequency}
  \label{Table:wordFrequency}
  \centering
  \begin{tabular}{lll}
    \toprule
    Most Frequent & Least Frequent\\
    \midrule
    pain & prepped \\
    pt & colonoscopy \\
    gout & eval \\
    day & mvc  \\
    swelling & femur  \\
    \bottomrule
  \end{tabular}
\end{table}

\subsubsection{ChatGPT}
GPT 4.0 is a large multimodal model that processes both image and text inputs to generate text outputs. 
It has been widely studied for its potential applications in dialogue systems, text summarization, and machine translation.
As reported by OpenAI, it highlights GPT 4.0's performance in various complex scenarios, notably on a simulated bar exam, where it scored in the top 10\%, contrasting with GPT 3.5's bottom 10\% score \cite{achiam2023gpt}. 
Due to this, we choose GPT 4.0 for our classification task over GPT 3.5.
ChatGPT is trained through a two-step process: pre-training and fine-tuning with Reinforcement Learning from Human Feedback (RLHF). 
In pre-training, the model learns general language patterns from a large dataset of publicly available text. 
Fine-tuning with RLHF refines its responses, using human feedback to guide the model toward producing outputs that align with human-like behavior and expectations \cite{achiam2023gpt}.
Despite the limitations of GPT 4.0 such as hallucinations, limited context windows, and lack of learning from experience, it represents a significant advancement in language processing capabilities.

\begin{figure*}[tbh]
\centerline{\includegraphics[width=0.95\textwidth]{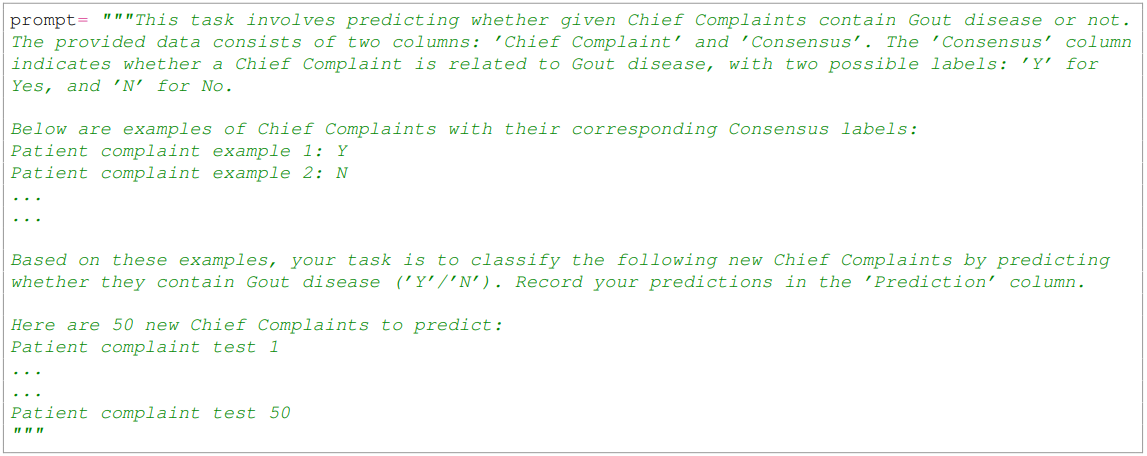}}
\caption{A Sample Prompt for 2-Class Prediction}
\label{Fig:prompt2class}
\end{figure*}

\subsubsection{Claude}
Claude AI, developed by Anthropic, is a chatbot designed for text generation and conversation. 
Claude 3 offers three models—Opus, Sonnet, and Haiku—each with varying capabilities. 
Haiku, the high-speed version for business, processes text three times faster than its peers, handling up to 21,000 tokens per second for prompts under 32,000 tokens \cite{anthropicClaudeHaiku}.
Opus, the paid version, designed for longer contexts, can process up to 100,000 words, making it ideal for complex tasks. 
Sonnet, the free version, has a smaller token limit but is more cost-effective, and suitable for smaller projects \cite{anthropicMeasuringPersuasiveness}. 
In our experiments, we evaluate the performance of disease identification using Claude 3 Opus, which is claimed to be the most intelligent AI model out of these three available versions.

\subsubsection{Gemini}
Gemini is a family of advanced multimodal models developed by Google that demonstrate notable capabilities across image, audio, video, and text understanding. 
The Gemini family comprises three versions: Ultra, Pro, and Nano, each tailored to specific application needs. 
Gemini Ultra, the most capable model, achieves state-of-the-art performance in 30 out of 32 benchmarks, excelling in reasoning, image understanding, video comprehension, and audio processing. 
Pro offers a balance between performance and efficiency, designed for scalable deployment on Google’s Tensor Processing Units (TPUs). 
Nano, designed for on-device applications, is the most efficient, with two sub-versions, Nano-1 with 1.8 billion parameters and Nano-2 with 3.25 billion parameters, optimized for low- and high-memory devices, respectively \cite{team2023gemini}. 
Gemini models are built on the transformer architecture with enhancements for stability and scalability, supporting a context length of 32k and multi-query attention for optimized inference. 
These models can process textual input interleaved with various media, including natural images, charts, PDFs, videos, and audio, and they can output text and image data. We select Gemini Ultra 1.0 as part of our disease prediction task.

\subsection{BERT}

In contrast to the LLM-based AI chatbots, we use a transformer-based architecture named Bidirectional Encoder Representations from Transformers (BERT) which is an NLP model that pretrains deep bidirectional representations from unlabeled text and is then fine-tuned with labeled text for specific NLP tasks \cite{devlin2018bert}. 
We select two variants of BERT, Clinical BERT, and BERT Base Uncased, to predict gout disease by inferring from patient complaints \cite{alsentzer2019publicly}.
We convert the categorical labels into numeric values using a label encoder, then tokenize the text data with the Clinical BERT and BERT base uncased tokenizer. 
This tokenization includes handling attention masks and applying truncation for uniformity. 
These models are fine-tuned with a pre-trained configuration, using an AdamW optimizer and varying learning rate scheduler for optimization. 
To account for class imbalance, we apply class weights in the cross-entropy loss function.

\section{Result Analysis}
We conduct a thorough analysis to evaluate the performance of different AI chatbots on the CC dataset. 
We aim to understand how well these AI chatbots could classify the CC data related to gout disease under varying conditions, such as different training levels. 
To do this, we train each chatbot with a specific number of examples, following the few-shot approach \cite{brown2020language}, and then examine them on the randomly selected 50 test patient complaints. Figure~\ref{Fig:prompt2class} demonstrates a sample prompt prepared for the 2-Class prediction task. 
We calculate F1 scores, which reflect the balance between precision and recall, to gauge the accuracy of the chatbots' predictions. 
We opt for the F1 score over accuracy, precision, and recall due to its emphasis on both precision and recall, providing a balanced measure in contexts where class imbalance might skew accuracy. 
The F1 score is particularly valuable when both false positives and false negatives carry significant implications, ensuring a holistic view of model performance \cite{goutte2005probabilistic}. 
This choice supports a more accurate evaluation in scenarios like ours, where accurate disease detection is critical. Equation~\ref{Eq:F1score} reflects the balance between precision and recall to calculate the F1 score.

\begin{equation}
F_1 = 2 \times \frac{{\text{Precision} \times \text{Recall}}}{{\text{Precision} + \text{Recall}}}
\label{Eq:F1score}
\end{equation}

\begin{figure*}[tb]
\centerline{\includegraphics[width=0.95\textwidth]{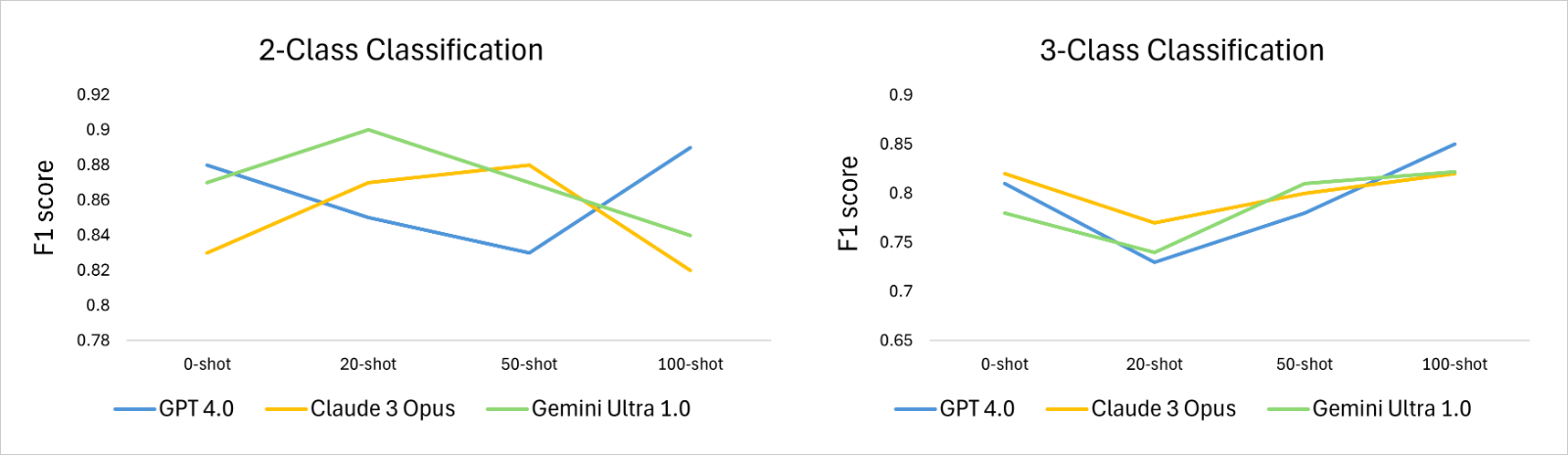}}
\caption{F1 Score Analysis of Different AI Chatbots}
\label{Fig:F1_score_comparison}
\end{figure*}

Initially, we filtered the complaints to include only those with labels indicating two classes, either `Y' or `N'. In the two-class scenario, GPT 4.0 and Gemini Ultra 1.0 demonstrated strong performance, with F1 scores that generally increased with more training examples. GPT 4.0 had its peak at 100-shot learning with an F1 score of 0.89, suggesting that it excels with more training data. In contrast, Gemini Ultra 1.0 achieved its best result at the 20-shot learning (0.90), indicating strong adaptability with fewer examples. Claude 3 Opus showed its peak performance at the 50-shot learning (0.88), with a dip in the 100-shot scenario, suggesting a possible plateau or overfitting. Overall, GPT 4.0 showed steady improvement with more training data, while Gemini Ultra 1.0 exhibited strong performance with minimal training.

For the three-class predictions (`Y'/`N'/`U'), GPT 4.0 again exhibited a steady increase in performance, with its peak at the 100-shot learning (0.85). Claude 3 Opus remained relatively consistent, with F1 scores ranging between 0.77 and 0.82 across all shot levels, indicating that it maintains stable performance with varying training examples. Gemini Ultra 1.0 achieved its best performance at the 50-shot learning (0.81), with a slight decline at the 100-shot learning, indicating a possible saturation point. These results suggest that while GPT 4.0 tends to improve with more training data, Claude 3 Opus remains steady, and Gemini Ultra 1.0 reaches its optimal performance with fewer examples. Figure~\ref{Fig:F1_score_comparison} exhibits the statistical performance of different AI chatbots for 2-class and 3-class classification setting.

While comparing the two-class and three-class scenarios, we found that GPT 4.0 consistently improved with more training data in both cases, indicating strong adaptability to different class structures. Claude 3 Opus maintained steady performance across both scenarios, suggesting robustness in handling varied classes. Gemini Ultra 1.0 showed better performance in the two-class scenario with fewer examples, but its performance in the three-class scenario was less consistent, indicating that it may struggle with increased complexity. Overall, GPT 4.0 seemed to adapt well to additional training data, while Claude 3 Opus remained stable across different class structures, and Gemini Ultra 1.0 performed better with simpler class structures and fewer examples.

In our experiment, we extended our analysis to assess transformer-based BERT's performance on this small dataset, contrasting its results with those of the AI chatbots. We selected Clinical BERT and BERT Base Uncased for disease classification. Clinical BERT's performance in the two-class scenario, with an F1 score of 0.68, indicated a moderate level of effectiveness in handling simpler classifications. However, in the three-class scenario, its F1 score dropped to 0.36. This could be attributed to Clinical BERT's specialization in medical terminology, which might have hindered its effectiveness with a dataset based primarily on patient feedback containing more generic words instead of medical jargon. Additionally, BERT Base Uncased showed a different trend, achieving an F1 score of 0.47 in the two-class scenario and 0.43 in the three-class scenario, demonstrating a more consistent but lower overall performance.

\begin{table}[tbp]
\centering
\caption{Peak Performance Across All Scenarios}
\label{Table:PeakPerfromance}
\begin{tabular}{*5l}
\toprule
Model & \multicolumn{2}{c}{F1 score} & \multicolumn{2}{c}{Accuracy}\\
\cmidrule(r){2-3}
\cmidrule(r){4-5}
{}   & 2-Class   & 3-Class    & 2-Class   & 3-Class\\
\midrule
BERT & 0.47 &  0.43 & 0.55  & 0.54 \\
ClinicalBERT & 0.68 &  0.36 & 0.67  & 0.45 \\
GPT 4.0 & 0.89 & \textbf{0.85} & \textbf{0.91} & 0.82 \\
Claude 3 Opus & 0.88 &  0.83 & 0.88  & \textbf{0.87} \\
Gemini Ultra 1.0 & \textbf{0.90} &  0.82 & 0.90  & 0.81 \\
\bottomrule
\end{tabular}
\end{table}

Table~\ref{Table:PeakPerfromance} showcases the peak performance of all our selected approaches, with the bold-face values highlighting the top performers in each category.
We observed that Gemini Ultra 1.0 exhibited the best F1 score for the two-class scenario, indicating it has strong precision and recall when dealing with simpler classifications. However, in the three-class scenario, GPT 4.0 achieved the highest F1 score, suggesting it adapts well to increased complexity. In terms of accuracy, GPT 4.0 and Gemini Ultra 1.0 both generated the most reliable results across two-class and three-class scenarios, respectively, indicating robust performance with few-shot learning. At the lower end, we found that ClinicalBERT and BERT generally yielded lower F1 scores and accuracy, suggesting these models might require further tuning or large training corpora for disease prediction.

\section{Discussion}

AI chatbots are becoming increasingly prevalent in the medical domain, offering both opportunities and challenges for healthcare practitioners and patients. Even though it has shown promise in its ability to generate eloquent text outputs and simulate human-like interactions, the priority in healthcare is patient safety and accuracy, which can sometimes be at odds with the conversational abilities and creativity of AI models \cite{lee2023benefits}.
Our analysis indicates that AI chatbots demonstrate potential for disease classification from patient complaints, but their reliability varies significantly. Models like GPT 4.0 achieved relatively high accuracy scores, suggesting a promising ability to classify patient complaints accurately. However, these scores, even at their best, do not reach a level that ensures absolute reliability. 
One reason for that is AI chatbots, like OpenAI's ChatGPT, are trained on a broad dataset from the open internet, which makes them susceptible to reflecting web-based biases and associations \cite{fang2024bias}. Conversely, models specifically trained on biomedical corpus datasets, such as Galactica or PubMedGPT, might focus too much on biomedical publishing trends rather than real-world clinical data, leading to gaps in understanding patient-centric contexts \cite{au2023ai}.

In clinical contexts, chatbots must ensure the reliability of information, particularly when used as clinical decision-support tools by healthcare professionals. The key is to balance human-like interactivity with the critical need for precise and accurate medical information. Even with our best-performing accuracy score of 91\% for GPT 4.0, this accuracy level is still not reliable enough for critical medical data classification, emphasizing the need for rigorous validation and human oversight. Variations in performance across training levels and complexity, along with risks of biases inherent in AI training datasets, highlight the need for caution when using these models for critical healthcare applications. This variability suggests that while AI chatbots can aid disease prediction, they must be used alongside human expertise to ensure accuracy and patient safety.

\section{Conclusion}
AI chatbots leveraging LLMs like GPT 4.0, Claude 3 Opus, and Gemini Ultra 1.0 show potential in predicting diseases from patient complaints, but their reliability varies. Our study indicates that GPT 4.0 performs best with more training data, while Gemini Ultra 1.0 excels with fewer examples, and Claude 3 Opus maintains consistent performance. Despite their advancements, these models do not yet guarantee the accuracy required for critical medical decisions, with variability across training levels and complexity. The findings underscore the importance of human oversight and rigorous validation in AI-based healthcare applications. To ensure patient safety, AI chatbots should complement, not replace, human expertise. Further research and refinement are necessary to enhance their reliability for disease prediction in clinical settings.

\bibliographystyle{IEEEtran}
\bibliography{Ref}
\vspace{12pt}

\end{document}